\documentclass[12pt]{article}
\usepackage{graphicx}
\usepackage{indentfirst,csquotes}

\usepackage{amssymb,amsthm,amsmath}
\usepackage{xcolor,paralist,hyperref,titlesec,fancyhdr,etoolbox}
\usepackage{authblk}
\usepackage{orcidlink}
\usepackage{siunitx}

\title{Deep learning recognition and analysis of Volatile Organic Compounds based on experimental and synthetic infrared absorption spectra} 
\author[1,2]{Andrea Della Valle\,\orcidlink{0009-0005-7405-1032}}
\author[3]{Annalisa D'Arco\,\orcidlink{0000-0001-7990-5117}}
\author[3]{Tiziana Mancini\,\orcidlink{0000-0003-4399-3869}}
\author[4]{Rosanna Mosetti\,\orcidlink{0000-0001-6768-3967}}
\author[3]{Maria Chiara Paolozzi\,\orcidlink{0000-0002-3714-8303}}
\author[3]{Stefano Lupi\,\orcidlink{0000-0001-7002-337X}}
\author[2,5]{Sebastiano Pilati\,\orcidlink{0000-0002-4845-6299}}
\author[1]{Andrea Perali\,\orcidlink{0000-0002-4914-4975}}

\affil[1]{School of Pharmacy, Physics Unit, University of Camerino, Camerino (MC), Italy}
\affil[2]{INFN-Sezione di Perugia, Perugia, Italy}
\affil[3]{Department of Physics, Sapienza University of Rome, Rome, Italy}
\affil[4]{Department of Basic and Applied Sciences for Engineering (SBAI), Sapienza University of Rome, Rome, Italy}
\affil[5]{School of Science and Technology, University of Camerino, Camerino (MC), Italy}

\date{}
\begin{document}
\maketitle

\begin{abstract}
Volatile Organic Compounds (VOCs) are organic molecules that have low boiling points and therefore easily evaporate into the air.
They pose significant risks to human health, making their accurate detection the crux of efforts to monitor and minimize exposure.
Infrared (IR) spectroscopy enables the ultrasensitive detection at low-concentrations of VOCs in the atmosphere by measuring their IR absorption spectra. However, the complexity of the IR spectra limits the possibility to implement VOC recognition and quantification in real-time. 
While deep neural networks (NNs) are increasingly used for the recognition of complex data structures, they typically require massive datasets for the training phase.
Here, we create an experimental VOC dataset for nine different classes of compounds at various concentrations, using their IR absorption spectra. 
To further increase the amount of spectra and their diversity in term of VOC concentration, we augment the experimental dataset with synthetic spectra created via conditional generative NNs.

This allows us to train robust discriminative NNs, able to reliably identify the nine VOCs, as well as to precisely predict their concentrations.
The trained NN is suitable to be incorporated into sensing devices for VOCs recognition and analysis.

\end{abstract}

\section{Introduction}

Today, monitoring indoor and outdoor air quality represents an increasingly urgent challenge. Specific actions are essential to mitigate exposure to biological and chemical pollutants, such as bacteria, viruses, and/or other organic compounds with toxic effects. At the same time, ongoing technological development is continually required to push detection methods beyond their limits, enabling targeted interventions to reduce emissions.
In this context, Volatile Organic Compounds (VOCs) represent one of the main hazards. VOCs are organic molecules produced by various natural processes, such as human and animal metabolic activities, and by anthropogenic processes, such as chemical and petrochemical industrial activities, construction materials, plastic production, cosmetics, or cleaning materials. As a result of their low boiling points, VOCs are therefore present in air in a wide range of environments \cite{Sicard2023, Salthammer2016Feb, Lanyon2005, Rizk2018Nov, Galstyan2021}.
Exposure to VOCs has been correlated with various health conditions \cite{Li2021Jan}, such as leukemia \cite{Lamm2009Dec}, neurocognitive disorders \cite{Mazzatenta2015Apr} or cancer \cite{Talibov2018Aug}.
However, the devices and instrumentation currently used to monitor and identify VOCs have intrinsic limitations related to the gas sampling and detection processes \cite{Galstyan2021}. The most critical limitations are weak sensitivity at low detectable concentrations, long response times that are not suitable for real-time implementation, and virtually no chemical specificity. Furthermore, many of these systems are not equipped to recognize or quantify multiple VOCs simultaneously, which is an important limitation, especially considering that current legislation does not adequately address co-exposure scenarios \cite{leggeeuropea, gazzettaufficiale}.
To address these limitations, monitoring devices with high sensitivity, capable of detecting very low concentrations of analytes, enabling real-time monitoring, and allowing easy chemical discrimination among molecules, are highly desirable. Achieving such performance requires, among other endeavors, the development of advanced numerical analysis algorithms. This is one of the core tasks addressed in this article. Specifically, we focus on the realization of machine learning (ML) methods for the  identification and quantification of VOCs through their infrared (IR) absorption spectra.

Vibrational spectroscopies, including IR and Terahertz (THz) spectroscopies, offer the potential for highly sensitive detection of VOCs at very low concentrations and with a direct and reliable discrimination \cite{Galstyan2021, Radica2021, darco2022, DArco2022a}. In IR spectroscopy, this is due to the ability to identify and resolve numerous molecular absorption lines. This allows  selective and sensitive identification of various VOCs at ambient concentration levels, typically in the part per million (ppm) range \cite{Radica2021, darco2022}. This is particularly advantageous for detecting common air pollutants that have strong IR absorptions. However, IR absorption spectra of VOCs often exhibit complex spectral features, including overlapping absorption peaks associated with different chemical bonds and vibrational modes, along with confounding background noise. These aspects make accurate identification and analysis particularly challenging. In fact, ML is increasingly applied to the management and analysis of spectral data \cite{mishra2022, ho2019, riad2019, wang2021, VOCNet2022, Chowdhury2024Mar, Chowdhury2025Mar}, suggesting that it might also be suitable for the identification of VOCs.
For example, convolutional neural networks (CNNs) have been used to predict the concentration of specific substances, but without performing automatic identification, thus limiting their application to cases where the target substance is known a priori \cite{Tian2021Sep}.
Training deep neural networks (NNs) typically requires vast amounts of data, which poses a major challenge for implementing NN-based VOC recognition algorithms. In this work, we address this issue through a two-pronged strategy: (i) conducting experiments to acquire VOC spectra and (ii) producing synthetic spectra using generative NNs \cite{davaslioglu2018, chung2024, schiemer2024}.

First, we collect a substantial experimental dataset of VOCs via Fourier-Transform IR (FTIR) spectroscopy. The dataset contains nine classes of different compounds, each analyzed at various concentrations using a Vertex 70v Michelson interferometer equipped with a multipass gas cell \cite{Radica2021, darco2022}. 
Based on this dataset, we train a first set of discriminative NNs that, thanks to a tailored two-head design, are capable of classifying spectra based on the presence of any of the nine VOCs, as well as  predicting their concentrations. 
Most relevantly, we demonstrate through the saliency map how our CNN automatically extract a larger amount of features from the spectra with respect to a manual analysis. 
To augment the experimental dataset, we implement a conditional variational autoencoder (CVAE) and train it  to generate synthetic VOC absorption spectra featuring the same characteristic as the experimental data \cite{Kingma2022, higgins2018, Sohn2015}. Notably, our CVAE is designed to allow controlling the VOC class and its concentration.

Our statistical analysis shows that these synthetic spectra are as effective as the experimental data for the training of discriminative NNs. Thus, they are suitable for expanding the size and balance of the training dataset, including also VOC concentrations not available in the experimental acquisition.
As we demonstrate, training on the augmented datasets improves the robustness of the discriminative NNs, allowing them to accurately identify VOCs and predict their concentrations under broader concentration conditions.

These findings set the basis for future incorporation of automatic VOC recognition into sensing devices. 
To favor further developments in this direction, we provide a vast database of synthetic VOC absorption spectra at the open Repository of Ref.~\cite{DellaValle2025Dec}.

The remainder of the article is organized as follows: 
the experimental and ML protocols are described in Section \ref{secmaterials};
the performance of our ML algorithms, including both generative and discriminative NNs, is tested in Section~\ref{secresults};
in Section~\ref{secconclusions} we summarize our main findings and discuss relevant future research directions.

\section{Materials and Methods}
\label{secmaterials}

\subsection{Experimental acquisition of spectra}

\begin{figure}[ht!]
    \centering
    \includegraphics[width=1\textwidth]{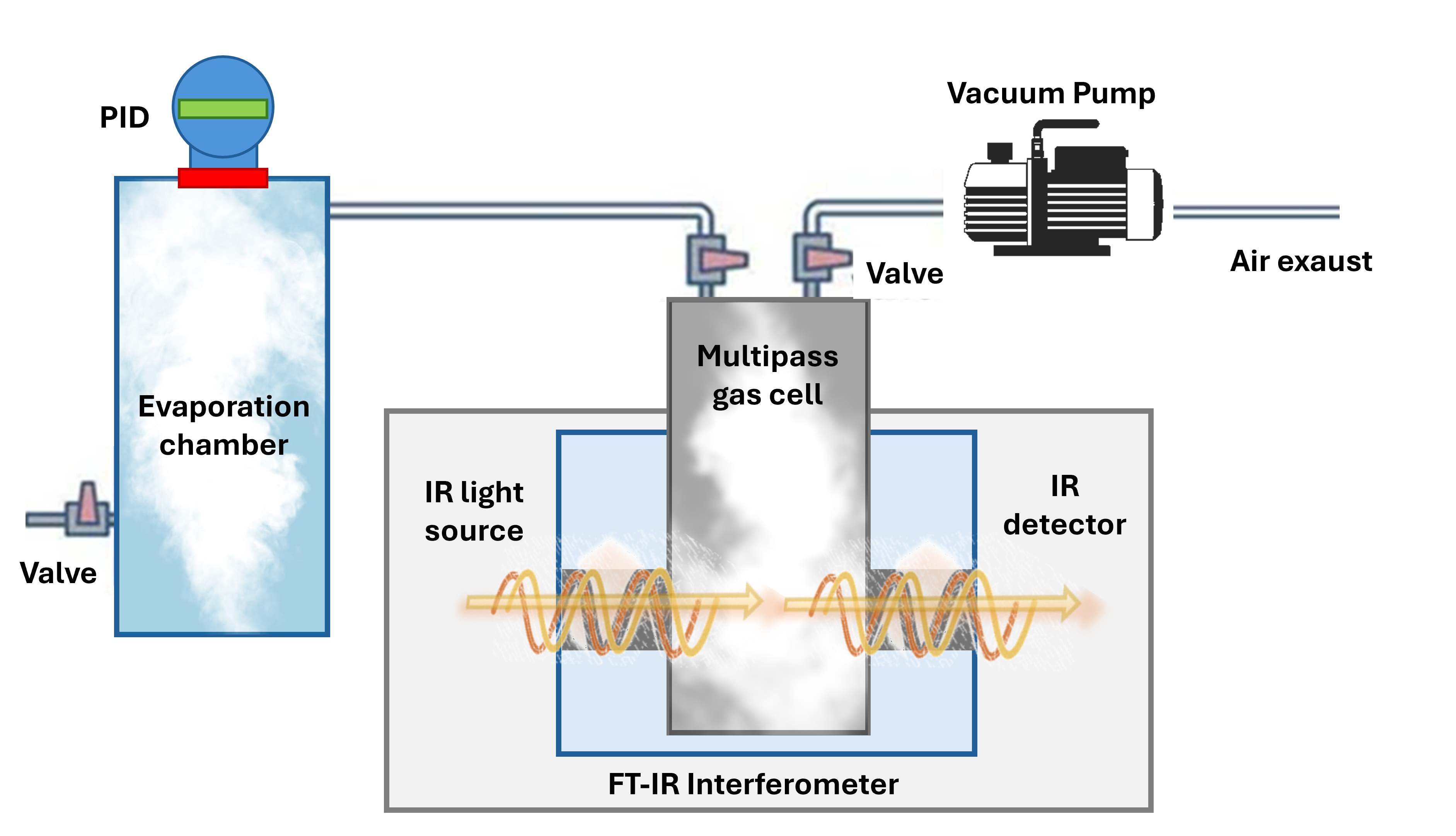}
    \caption{Schematic view of the setup used for collecting VOCs absorption spectra. Liquid VOCs are added in varying amounts to the evaporation chamber using a micropipette. Gaseous VOCs concentrations are continuously monitored via PID sensor controlled by LabVIEW$^{TM}$. Once the PID signal stabilizes, indicating vapor-liquid equilibrium, the valve to the pre-evacuated multipass spectrometer cell is opened. The cell, sealed with KBr windows and evacuated to a few mbar using an Edwards T-Station 85, allows rapid gas transfer. Spectral acquisition begins immediately upon valve opening, with PID readings recorded simultaneously.}
    \label{fig:setup}
\end{figure}

Here, we analyze nine gaseous VOCs: acetone (CAS 67-64-1, Sigma Aldrich), ethanol (CAS 64-17-5, Sigma Aldrich), isopropanol (CAS 67-63-0, Sigma Aldrich), styrene (CAS 100-42-5, Carlo Erba Reagents), benzene (CAS 71-43-2, Sigma Aldrich), toluene (CAS 108-88-3,Sigma Aldrich), ortho-xylene (CAS 95-47-6, Sigma Aldrich), para-xylene (CAS 106-42-3, Sigma Aldrich), and meta-xylene (CAS 108-38-3,Sigma Aldrich) with purity $\geq 99\%$.
These spectra are collected using a Bruker Vertex 70V interferometer equipped with a GEMINI Mars Series multipass gas cell $2 \mathrm{L}$/$10 \mathrm{m}$ with a nominal path length of $10 \mathrm{m}$ and a volume of $2 \mathrm{L}$ \cite{darco2022}. Each spectrum is the average of 64 scans in the $400-5000 \mathrm{cm}^{-1}$ frequency range, collected with a spectral resolution of $2 \mathrm{cm}^{-1}$.
A schematic of the experimental setup, along with a detailed description of the measurement collection procedure, is provided in Fig.~\ref{fig:setup}.

In summary, high sensitivity towards low concentrations is achieved in the IR measurements using a multipass gas cell
connected to a sealed evaporation chamber. The evaporation chamber is equipped with a commercial Photo-Ionization Detector (PID) TA-2100 Styrene Detector (Mil-Ram Technology, Inc., Fremont, CA, USA) for real-time monitoring of evaporated VOCs. The PID is calibrated to detect styrene concentrations ranging from 1 to 100 parts per million (ppm) with a sensitivity of 1 ppm.
VOC concentrations in the multipass gas cell is estimated based on the PID readings of the concentrations in the volume of the expansion chamber ($0.6 \mathrm{L}$). When the expansion chamber is connected to the multipass gas cell, the gas expands in the new volume $(V_1+V_2)$, and its concentration and error are 
\begin{align}
C &= \mathrm{PPM} \cdot CF \cdot \frac{V_1}{V_1 + V_2}
\\
\rm{err}(C) &= CF \cdot \frac{V_1}{V_1 + V_2}
\end{align}
where $\mathrm{PPM}$ is the PID reading at the beginning of the experiment, $V_1$ is the volume of the evaporation chamber $0.6 \mathrm{L}$, $V_2$ is the volume of the multipass gas cell $2 \mathrm{L}$, and $CF$ is the factor used to convert the concentration provided by the PID and referred to styrene to the concentration of other VOCs. In particular, $CF$ values are $2.75$ for Acetone, $1.325$ for Benzene, $30$ for Ethanol, $15$ for Isopropanol, $1.1$ for m-Xylene, $1.15$ for o-Xylene, $0.975$ for p-Xylene, $1$ for Styrene, and $1.25$ for Toluene.

\subsection{Dataset}
\begin{figure}[ht!]
    \centering
    \includegraphics[width=1\textwidth]{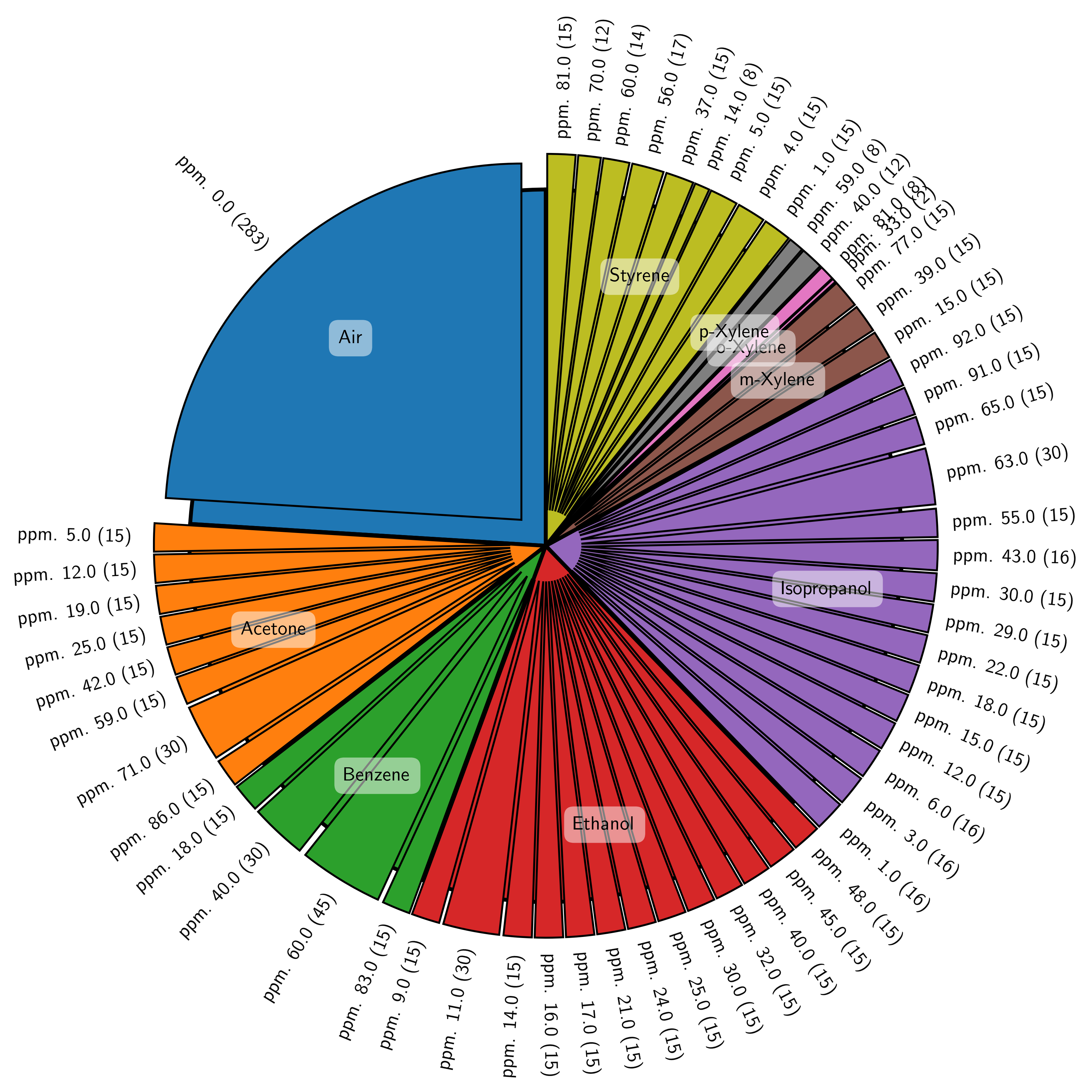}
    \caption{Experimental dataset composition, divided by class and concentration (in brackets the number of IR spectra for each condition). The dataset is composed of $1253$ IR spectra. }
    \label{fig:composition-dataset}
\end{figure}

The experimental dataset is composed of $1230$ IR absorption spectra, divided into nine groups of pure VOCs, specifically \emph{acetone}, \emph{benzene}, \emph{ethanol}, \emph{isopropanol}, \emph{xylenes (orto-, meta-, para-)}, \emph{styrene}, and \emph{toluene}, and also a class named \emph{air}, which includes IR spectra collected in the absence of any VOC within the evaporation chamber.
The frequency range used for the analysis is $\SIrange{700}{1300}{\cm^{-1}}$ interpolated into $622$ channels. This range is chosen due to the abundance of spectral features representative of the compounds under investigation and the absence of specific peaks of $CO_2$ or water vapor within these frequencies. 
The dataset is unbalanced in terms of VOC classes, as shown in Fig.~\ref{fig:composition-dataset}.
Yet, the overall number of spectra significantly extends previously available datasets~\cite{darco2022, Mancini2025Jun}.
The issue of under-representation of specific molecular classes can be addressed through data augmentation, as discussed below.

\subsection{Machine learning methods}
The computational part of this work involves three major tasks:
\begin{itemize}
\item[i)] training a discriminative NN --  we refer to it as the \emph{basic} model -- on the experimental dataset to identify VOCs and to predict their concentration from the input IR spectra.
\item[ii)] Training a conditional generative NN  to produce synthetic spectra corresponding to chosen VOC classes and concentrations.
\item[iii)] Training two additional discriminative NNs, which we refer to as the \emph{enhanced} models, on datasets augmented either adding synthetic spectra (\emph{synthetic enhanced} model) or using a more conventional oversampling approach (\emph{oversample enhanced} model); their performances are compared against the \emph{basic} NN.
\end{itemize}
In the following, these three ML tasks are described in detail.

\subsubsection{Basic model}
\begin{figure}[ht!]
    \centering
    \includegraphics[width=0.6\textwidth]{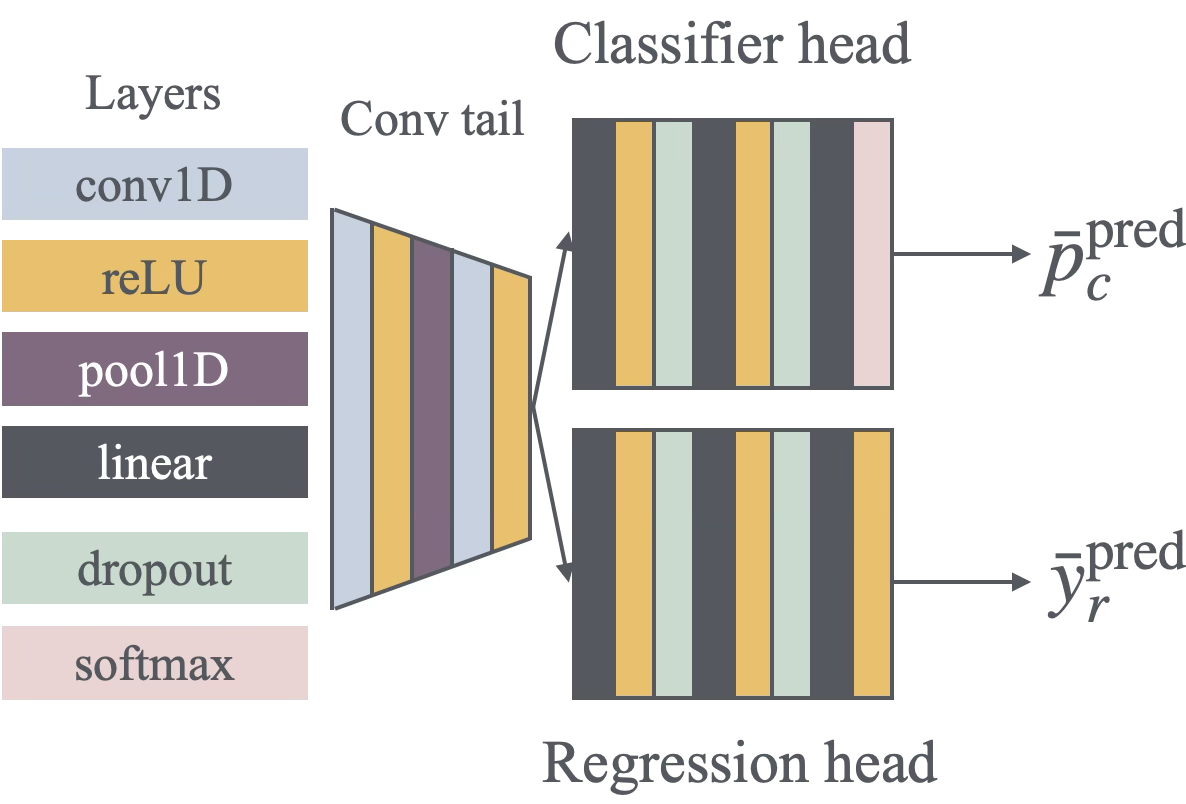}
    \caption{Architecture of the CNN used to predict the VOC class and concentration. The sequence of one-dimensional convolutional layers is followed by two separated MLPs: the classifier head and the regression head.}
    \label{fig:architectureMIX}
\end{figure}

For the discriminative NNs, we implement a tailored CNN, designed to predict both the VOC identity and its concentration. 
A well-known key strength of CNNs is their ability to automatically extract relevant system features, avoiding the use of hand-crafted descriptors \cite{mishra2022}, and thus the risk of introducing human bias. 
The architecture of our CNN is sketched in Fig.~\ref{fig:architectureMIX}, while the relevant parameters are reported in Table~\ref{tab:model_params}.
In short, two one-dimensional convolutional layers, alternated with pooling layers, are followed by a two-head structure, including a classifier head and a regression head. Both heads consist of a standard multi-layer perceptron (MLP) featuring two linear layers. The VOC class is represented via one-hot encoding, while the concentration is predicted via a multiple-output layer with nonzero concentration only at the index corresponding to the correct class. 
The activation function are the rectified linear unit (reLU) and the softmax function for the classifier head output.
Training is performed by minimizing a composition of standard loss functions, namely, the sum of the mean square error (MSE) of the concentration prediction and the cross-entropy for the multi-label classification. 
The standard k-fold split with $5$ folds is used for NN training.
Specifically, four folds are used for coefficient optimization, while the remaining fold represents the validation set used for hyperparameter selection. Five NNs are trained for each numerical experiment, rotating the validation fold. This allows us to assess the statistical fluctuations in the performance analysis. Specifically, error bars on performance scores are defined as the average standard error.

\begin{table}
    \centering
    \begin{tabular}{c|c}
       Layer parameter  & Value \\
       \hline
      No. channels in convolutional layers & (3, 3)  \\
      Convolutional kernel size   & 3  \\
      Average pooling kernel size  & 2  \\
      No. of hidden linear output features   & (256, 64) \\
      Dropout value & 0.5
    \end{tabular}
    \caption{Parameters of the discriminative NNs. The two output heads feature the same parameters.}
    \label{tab:model_params}
\end{table}

\subsubsection{Generative model}
\begin{figure}[ht!]
    \centering      \includegraphics[width=1\textwidth]{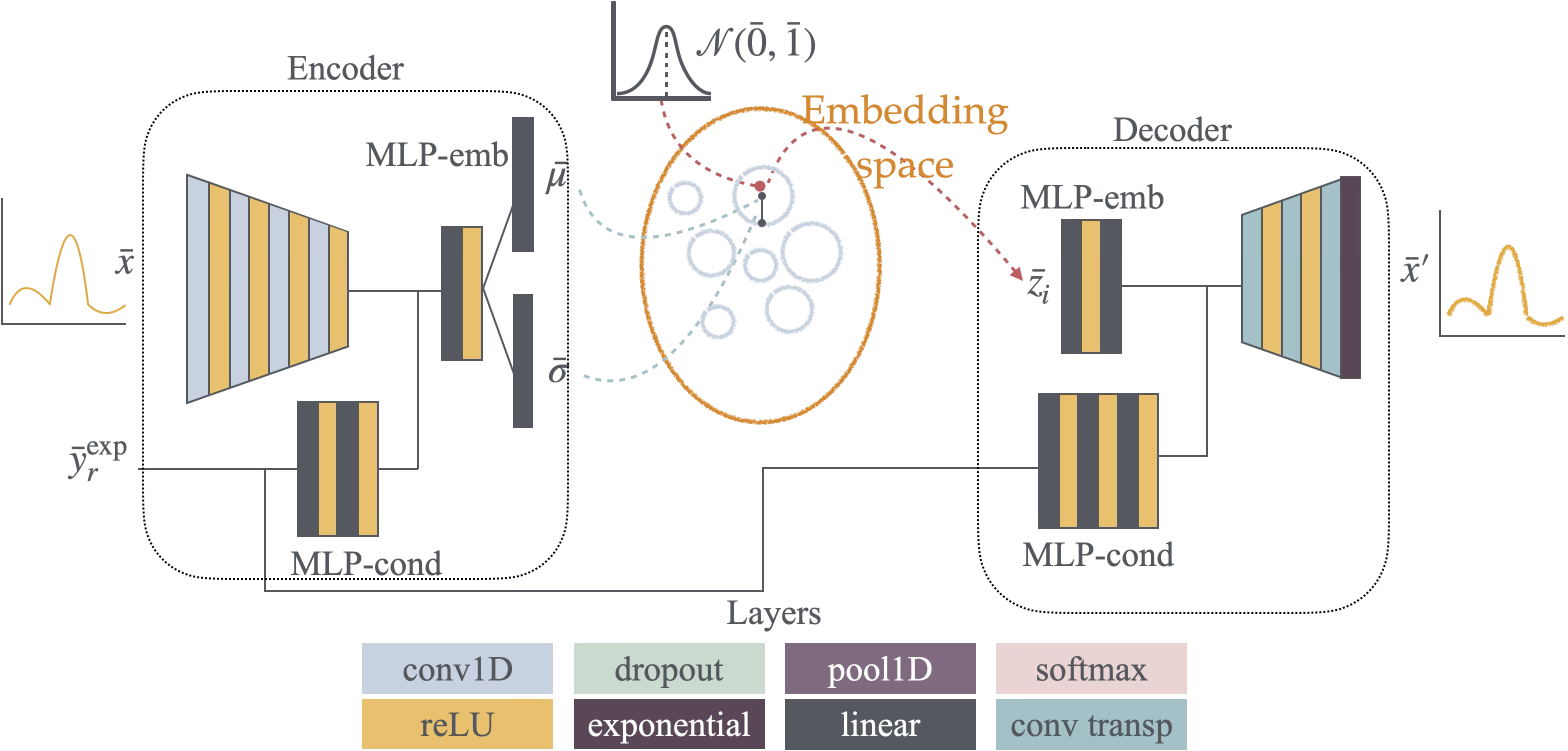}
    \caption{Architecture of the CVAE. It is composed on two main blocks: conditional encoder and decoder. After the training, the conditional decoder is used to generate spectra with desired conditions.}
    \label{fig:vae_training}
\end{figure}

To generate synthetic spectra that resemble the experimental data we adopt variational autoencoders (VAEs) \cite{Kingma2022, higgins2018}, modified to allow the generation to be conditioned using user-specified molecular classes and concentrations.
The architecture of standard VAEs consists of two main blocks: \emph{i)} an encoder, which maps the input data into a probabilistic compressed embedding space, and \emph{ii)} a decoder, which  reconstructs spectra from points in the embedded space. The conditioned variant extends the standard architecture by inserting the conditioning input -- in our implementation, a vector of VOC concentrations -- into both the encoder and the decoder. 
The conditioned encoder is composed of three main blocks of layers: \emph{i)} four convolutional layers, \emph{ii)} an MLP featuring two linear layers acting on the input condition (referred to as MLP-cond) and \emph{iii)} a final MLP featuring two linear layers (referred to as MLP-emb) processing the output of the other two blocks mapping the combined features on the embedding space.
The decoder is similarly composed of three main blocks: \emph{i)} an MLP featuring wo linear layers (referred to as MLP-emb) processing the embedding space, \emph{ii)} an MLP featuring two linear layers (referred to as MLP-cond) processing the input condition; their output are combined to construct a one-dimensional feature representation with 8 channels, processed by a final \emph{iii)} stack of transposed convolutional layers, followed by an exponential layer used to reconstruct the original spectra.
The activation function is reLU.
The architecture of our CVAE is sketched in Fig.~\ref{fig:vae_training}, and the main parameters are reported in Table~\ref{tab:encoder_decoder_params}.
\begin{table}
    \centering
    \begin{tabular}{c|c}
       Layer's parameter  & Value \\
       \hline
      No. channels in convolutional layers of encoder & (16, 16, 32, 32) \\
      Convolutional kernel size of encoder   & 3  \\
      No. output neurons in MLP-cond of encoder & (32, 64)  \\
      No. output neurons in MLP-emb of encoder   & (64, 16) \\
       \hline
      No. output neurons in MLP-cond of decoder & (32, 64, 77)  \\ 
      No. output neurons in MLP-emb of decoder  & (32, 77$\times$7) \\
      Kernel size in  transposed convolutional layers   & (5, 5, 4)  \\
      No. channels in transposed convolutional layers & (8, 8, 1) \\
      Stride of transposed convolutional layers & 2 \\
      Padding of transposed convolutional layers & 1 \\  
    \end{tabular}
    \caption{Parameters of the conditional encoder and decoder. }  \label{tab:encoder_decoder_params}
\end{table}
The CVAE is trained by minimizing the following loss function:
\begin{equation}
\mathcal{L} = \mathcal{L}^{\mathrm{recon}}(\bar{x}, \bar{x}') + D_{KL}(\mathcal{N}(\bar{\mu}, \bar{\nu}) || \mathcal{N}(\bar{0}, \bar{1})),
\label{eq:standardloss_cvae}
\end{equation}
where $\mathcal{L}^{\mathrm{recon}}(\bar{x}, \bar{x}')$ represents the mean squared error between the input $\bar{x}$ and output spectra $\bar{x}'$, and $D_{KL}(\mathcal{N}(\bar{\mu}, \bar{\nu}) || \mathcal{N}(\bar{0}, \bar{1}))$ is the Kullback-Leibler divergence between two multi-variate Gaussian distributions \cite{Kullback1951Mar}. The means $\bar{\mu}$ and variances $\bar{\nu}$ are predicted by the encoder. The second term in the loss function enforces regularization of the embedding space, allowing the generation of realistic spectra not only from points exactly corresponding to encoded spectra.

\subsubsection{Enhanced models}

For the \emph{enhanced} discriminative NNs, we adopt the same network architecture used for the \emph{basic} model. However, training is performed on augmented datasets.
\begin{figure}[ht!]
    \centering  
    \includegraphics[width=1\textwidth]{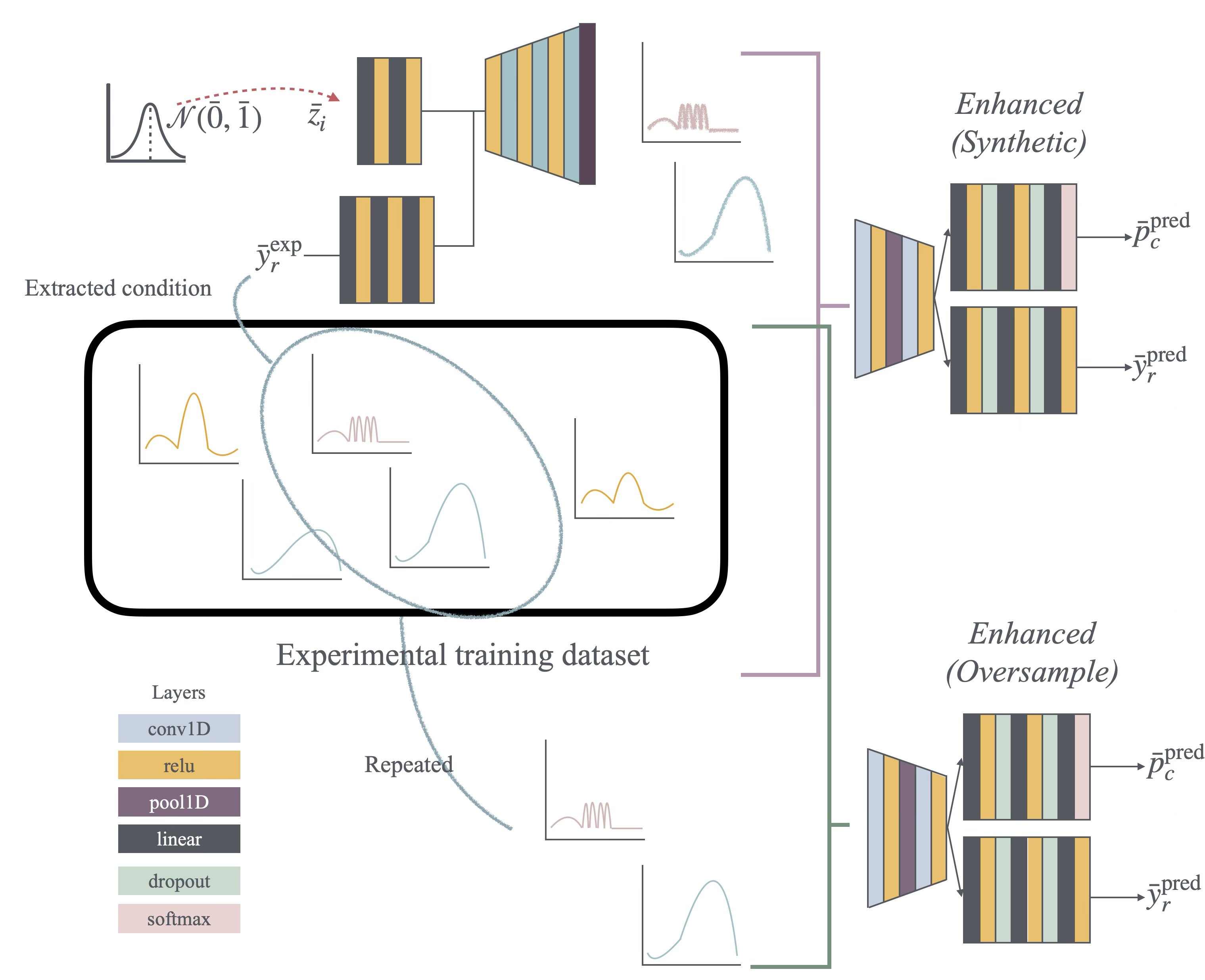}
    \caption{Training procedure for \emph{enhanced} discriminative NNs. }
    \label{fig:enhanced_training}
\end{figure}
In detail, in the case of the \emph{oversample enhanced} NN, at each training epoch a randomly selected subset of spectra for each class is repeated. 
In the \emph{synthetic} case, instead, the trained CVAE is used to generate the additional  data, with concentrations uniformly sampled for each class from the original training dataset. 
Different sizes of augmented training dataset were used to train separated \emph{enhanced} NNs, with each NNs trained on a dataset containing from 10 to 200 augmented spectra per class. The concentrations of the added spectra were randomized for each training epoch. The best model was selected by choosing the sample minimizing the average validation MSE over the 5 folds.
The training process for the two \emph{enhanced} models is sketched in Fig.~\ref{fig:enhanced_training}.

\subsection{Evaluation metrics}
To evaluate how accurately the discriminative NNs predict VOC concentrations, we determine the MSE and the coefficient of determination $R^2$.
Given a set of ground-truth target values $y_i^{\mathrm{true}}$ and predictions $y_i^{\mathrm{pred}}$ labeled with the index $i$, the $R^2$ score is defined as
\begin{equation}
R^2 =  1 - \frac{\sum_i (y_i^{\mathrm{true}} - y_i^{\mathrm{pred}}) ^2 }{\sum_i (y_i - \left<y\right>)^2 }, 
\end{equation}
where $\left<y\right>$ is the average of the ground-truth targets.
Notice that the score $R^2=1$ indicates perfect predictions, while the score $R^2=0$  is equivalent to a trivial model that predicts a constant equal to the average of the ground-truth values.
For the VOC-identification task, the performance is quantified by determining the  classification accuracy, namely, the ratio
$\mathrm{Acc}=N_{\mathrm{CP}}/N_{ \mathrm{total} }$ between the number of correct predictions $N_{\mathrm{CP}}$ and the number of instances $N_{ \mathrm{total} }$.
The metric uncertainties are reported as the standard error of the mean over the 5-fold repetitions.

\section{Results}
\label{secresults}

\subsection{Basic model}

\begin{figure}[ht!]
    \centering
    \includegraphics[width=0.8\textwidth]{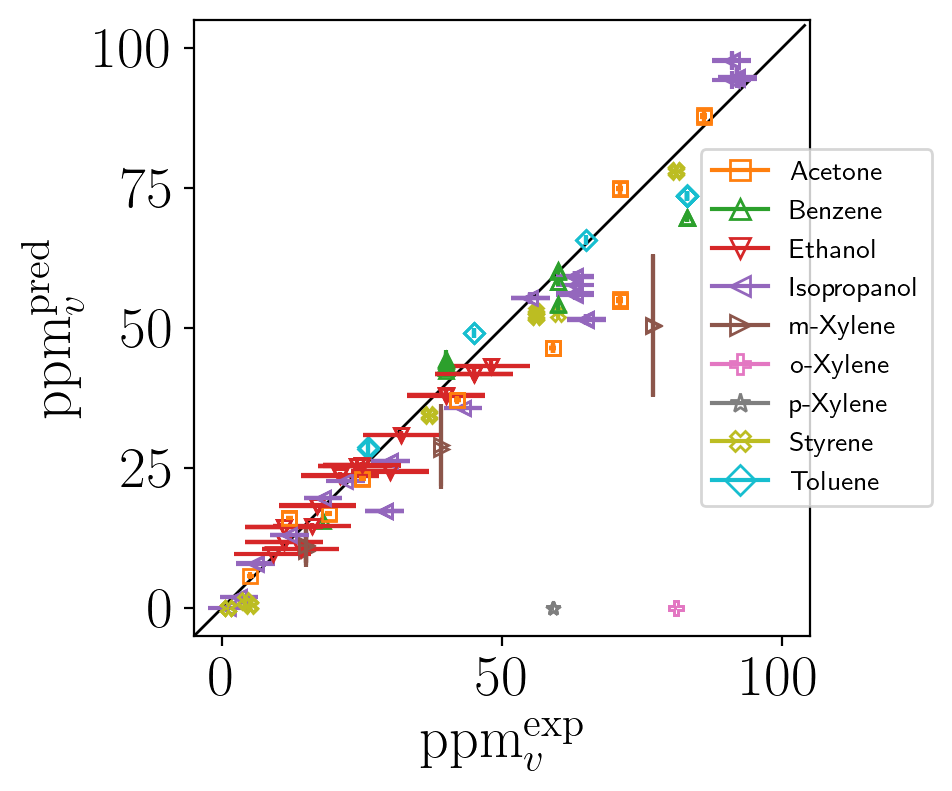}
    \caption{Concentrations predicted by the \emph{basic} NN \emph{vs.} ground-truth experimental values. The horizontal bars refer to the sensitivity of the PID. 
    }
    \label{fig:basic-scatterplot}
\end{figure}

\begin{figure}[ht!]
    \centering  
    \includegraphics[width=1\textwidth]{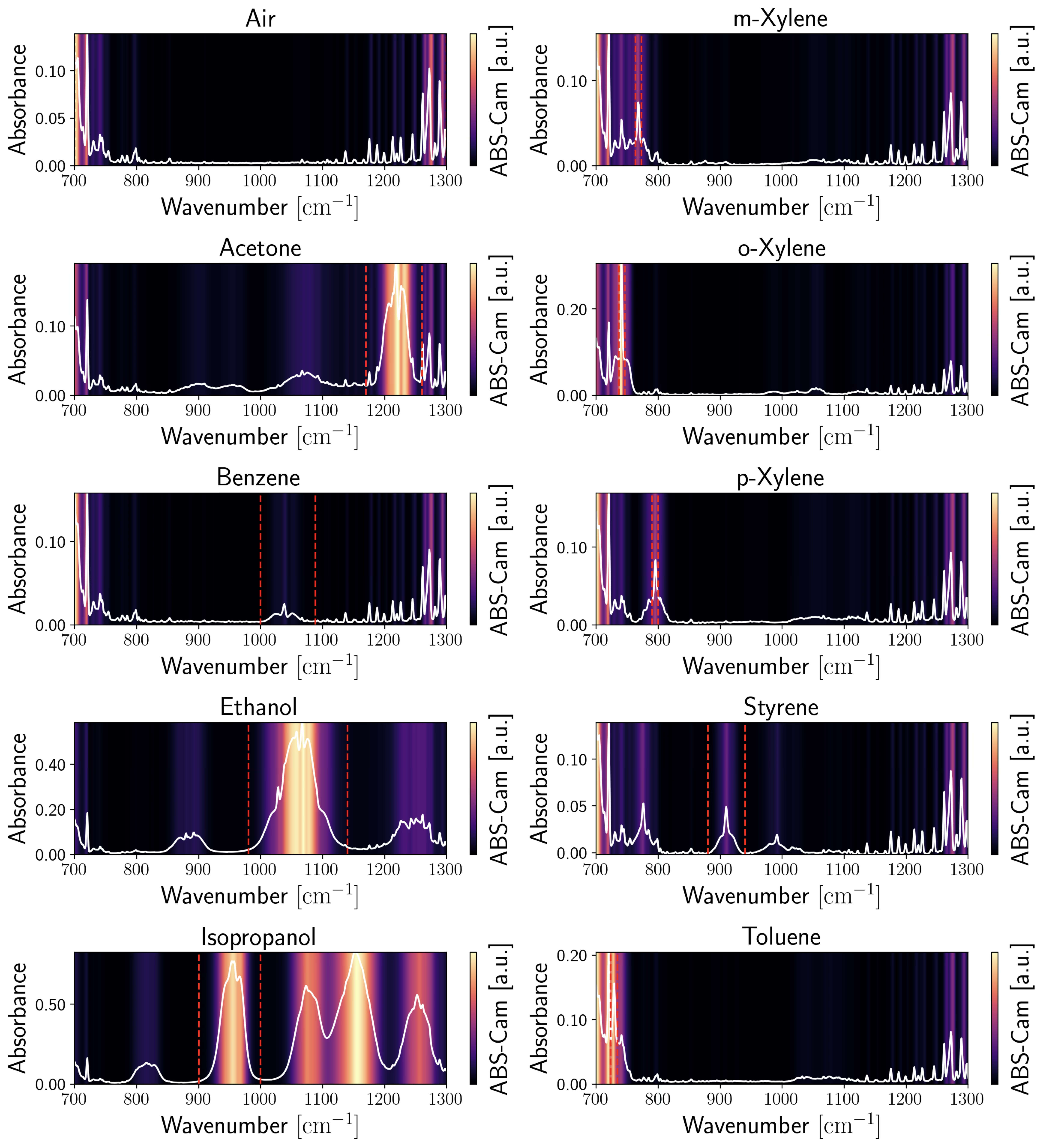}
    \caption{Abs-CAM analysis applied to representative spectra (white curves) for each VOC or air classes. The color intensity represents the relevance of each frequency channel (arbitrary units). The vertical dashed red lines bracket the manually selected frequency ranges \cite{Mancini2025Jun}.}
    \label{fig:ABS_cam}
\end{figure}

The performance of the \emph{basic} NN is analyzed in Fig.~\ref{fig:basic-scatterplot}.
While for most VOCs the concentration is predicted with fair accuracy, in the cases of ortho-xylene and meta-xylene, an erroneous vanishing concentration is predicted. We attribute this inconsistency to the scarcity of spectra in the corresponding classes, which highlights the importance of augmenting the experimental dataset.
For the VOC identification task, the \emph{basic} NN reaches a reasonable accuracy, namely $\mathrm{Acc}=99.5(4)\%$.
To gain some intuition about the functioning of our discriminative NN, we produce saliency maps via the Abs-CAM approach of Ref. \cite{Zeng2023Jun}, which we adapt introducing the generalized loss function described in Section \ref{secmaterials}. Saliency maps are based on information extracted from the last convolutional layer and provide insight into how the NN interprets the spectral features.
In Fig.~\ref{fig:ABS_cam}, we compare them with the spectral ranges (marked with red dotted lines) adopted in a previous study \cite{Mancini2025Jun} to train binary classifiers that identify the presence of VOCs. Quite importantly, for some classes, the saliency maps indicate that our CNNs leverage information outside the human-selected ranges, highlighting that manual feature selection may introduce biases and  potentially lead to the loss of useful information.
For instances, this is particularly clear in the case of \emph{Isopropanol}, where the saliency map highlights four to five spectral peaks instead of the single peak (in range $\SIrange{980}{1140}{\cm^{-1}}$) included in the human-based feature selection. Similar advantage can be found for \emph{Ethanol} and \emph{Styrene} as well. 

\subsection{Generative model}

\begin{figure}[ht!]
    \centering
    \includegraphics[width=1\textwidth]{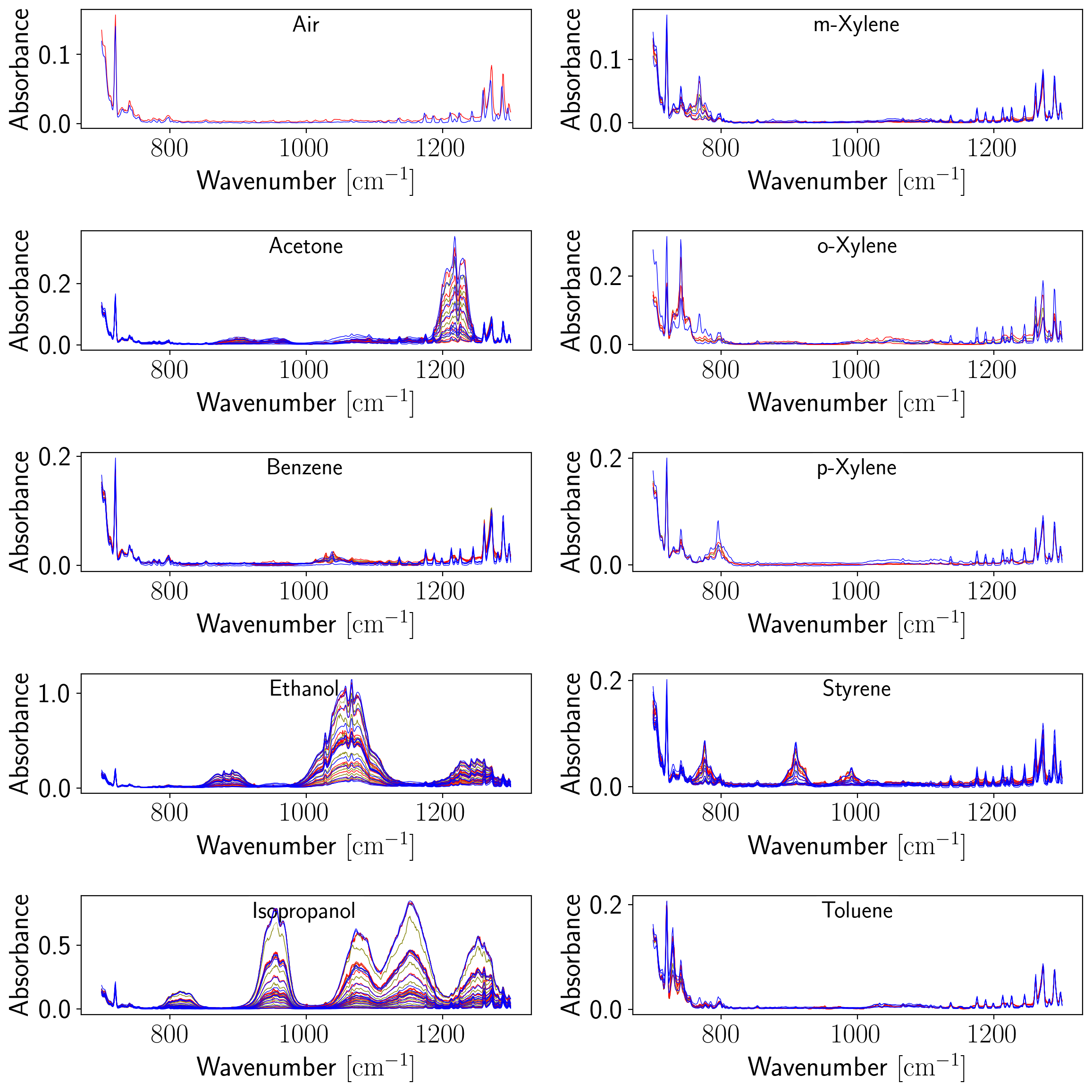}
    \caption{Comparison among representative experimental spectra (blues curves) and corresponding synthetic spectra generated by the CVAE (red curves) for concentrations randomly selected from the experimental dataset. Green curves represent synthetic spectra for intermediate concentrations not present in the experimental dataset.
    }
    \label{fig:generatedspectra}
\end{figure}

\begin{figure}[ht!]
    \centering
    \includegraphics[width=1\textwidth]{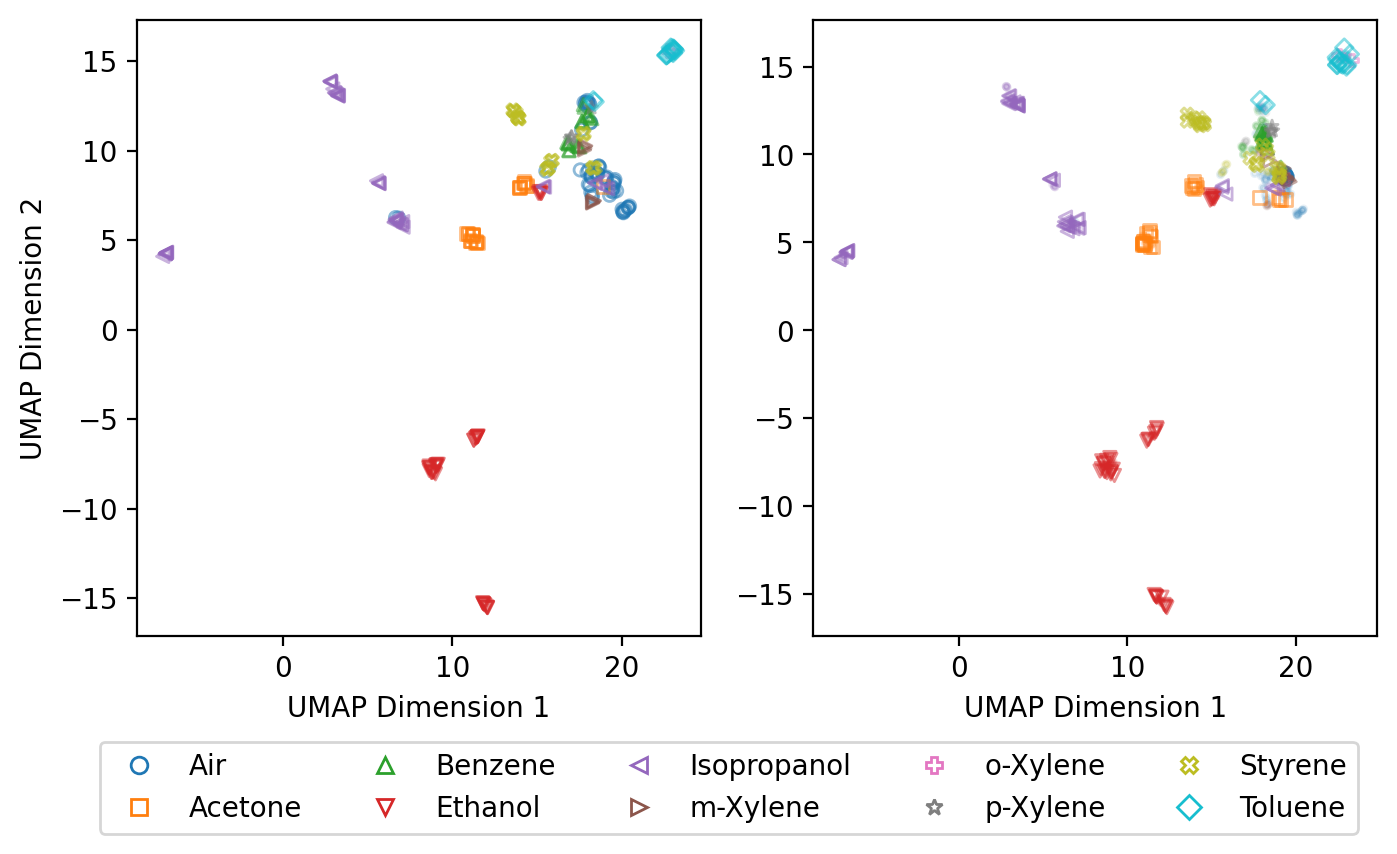}
    \caption{UMAP representation of experimental spectra (symbols in left panel and semi-transparent points in right panel) and synthetic spectra (symbols in the right panel). The concentrations used for generation match those of the experimental dataset.  
    }
    \label{fig:umap}
\end{figure}
In Fig.~\ref{fig:generatedspectra} we show examples of spectra generated using the CVAE for different VOCs. The visual comparison with the experimental data demonstrates a qualitative similarity.
A comparison is made via the UMAP \cite{McInnes2018Feb} representation, shown in Fig.~\ref{fig:umap}. 
The UMAP is implemented instead of PCA \cite{Hotelling1933AnalysisOA} since it preserves global structure of spectral data, providing more informative embeddings, as already analyzed in literature \cite{Bouzerda2025}.
The UMAP is trained on the experimental dataset, revealing that the spectra are organized in distinct clusters. Classes exhibiting distinctive spectral features are well separated in the UMAP space; however, others tend to overlap, particularly at low concentrations. This overlap is observable even in classes that are well separated at higher concentrations. 
Then, synthetic spectra with the same VOC concentrations are projected using the pre-trained UMAP. We find that they closely overlap with the experimental data, preserving their cluster organization. Again, this highlights the close resemblance between the experimental and synthetic spectra.

\subsection{Enhanced models}

\begin{figure}[h!]
    \centering
    \includegraphics[width=1\textwidth]{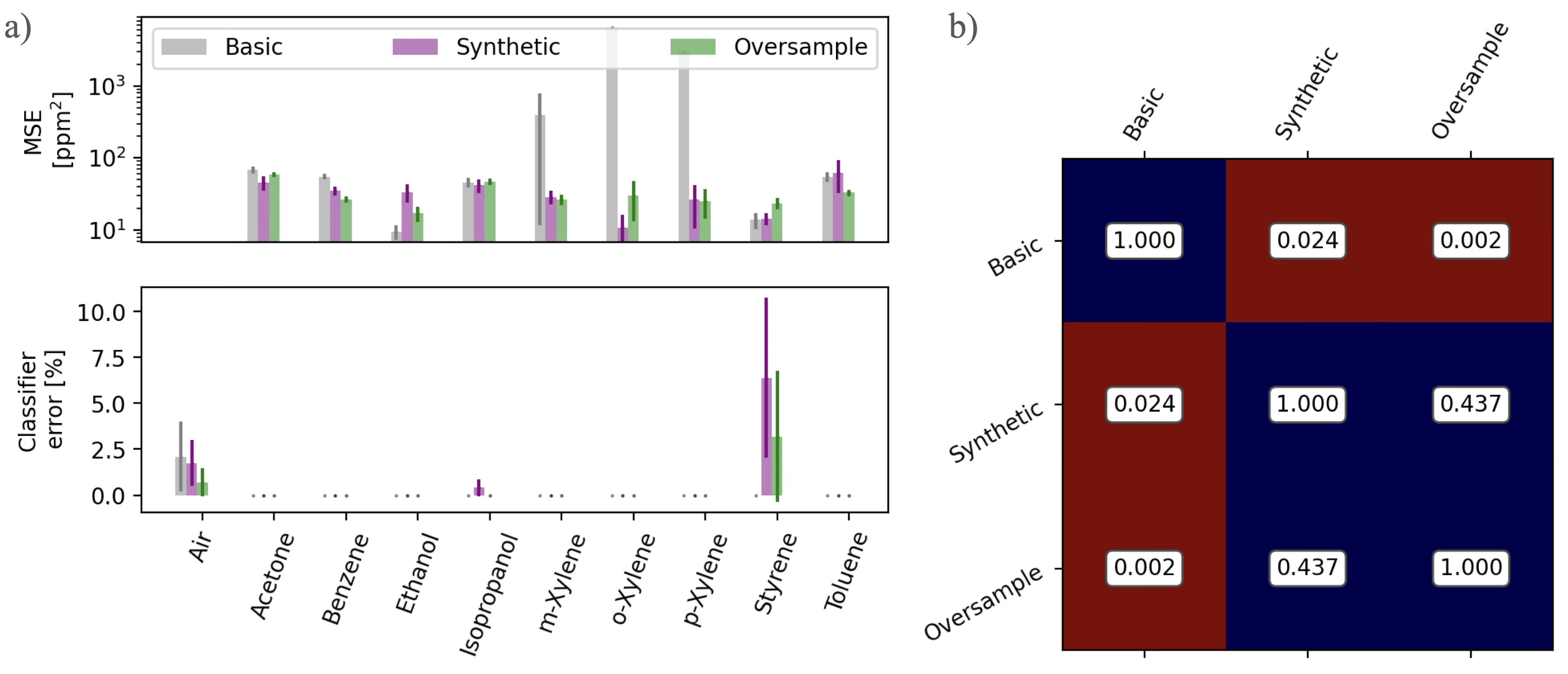}
    \caption{\textbf{(a)} Performance comparison in concentration prediction (upper panel) and in VOC class identification (lower panel) among the \emph{basic} NN and the \emph{synthetic} and \emph{oversample} enhanced models. \textbf{(b)} Dunn post-hoc analysis comparing MSE of the different NNs on test sets. Significance differences (set at $p < 0.05$) are indicated in red in the significance matrix, indicating that the \emph{oversample} and \emph{synthetic} NNs perform similarly, while both significantly outperform the \emph{basic} NN. }
    \label{fig:perfomance-enhanced}
\end{figure}

In Fig.~\ref{fig:perfomance-enhanced}(a) we compare the performance of the two \emph{enhanced} discriminative NNs against the \emph{basic} model. Both the \emph{synthetic} and \emph{oversample} enhanced NNs demonstrate a significant performance increase in concentration prediction, in particular for the underrepresented VOCs. 
The average MSE evaluated on the whole test set is  $142(7) \mathrm{ppm}^2$ for the \emph{basic} NN,  $29(1) \mathrm{ppm}^2$ for the \emph{synthetic} NN, and $26.7(3) \mathrm{ppm}^2$ for the \emph{oversample} NN.
The $R^2$ score computed separately for each class (cfr. Tables~\ref{tab:r2scores_enhancedgen}) shows an improvement of performance for both the \emph{enhanced} models, particularly in underrepresented classes.
\begin{table}
    \centering
    \begin{tabular}{c|c|c|c}
       VOC's class  & Basic  & Enhanced oversample & Enhanced synthetic \\
       \hline
      Acetone   & $0.972(3)$ & $0.975(2)$ & $0.980(4)$ \\
      Benzene   & $0.983(1)$ & $0.9913(9)$ & $0.988(1)$ \\
      Ethanol   & $0.986(3)$ & $0.975(5)$ & $0.95(1)$ \\
      Isopropanol   & $0.984(2)$ & $0.983(2)$ & $0.985(3)$ \\
      m-Xylene  & $0.8(2)$ & $0.985(2)$ & $0.983(3)$ \\
      o-Xylene  & $-0.008$ & $0.995(2)$ & $0.9983(7)$ \\
      p-Xylene  & $-0.02$ & $0.991(3)$ & $0.991(5)$ \\
      Styrene & $0.993(2)$ & $0.989(2)$ & $0.993(1)$ \\
      Toluene  & $0.988(2)$ & $0.9926(7)$ & $0.986(9)$ \\
    \end{tabular}
    \caption{$R^2$ scores for the three discriminative NNs. }
    \label{tab:r2scores_enhancedgen}
\end{table}
On the one hand, these findings highlight the importance of data augmentation; on the other hand, they indicate that synthetic spectra are essentially as effective as experimental data in the context of NN training.
To further analyze the behaviors of the three NNs, we perform a Kruskal comparison analysis, obtaining $F=9.98$, $p=0.0068$ \cite{Kruskal1952Dec, 2020SciPy-NMeth}, which underlies a statistical difference between the results of the three NNs. In detail, the Dunn post-hoc test \cite{Olive1964Aug, Terpilowski2019Apr} (see Fig.~\ref{fig:perfomance-enhanced}(b)) shows that there is no significant difference between \emph{oversample} and \emph{synthetic} NNs, and both differ significantly from the \emph{basic} NN.

\begin{figure}[h!]
    \centering
    \includegraphics[width=1\textwidth]{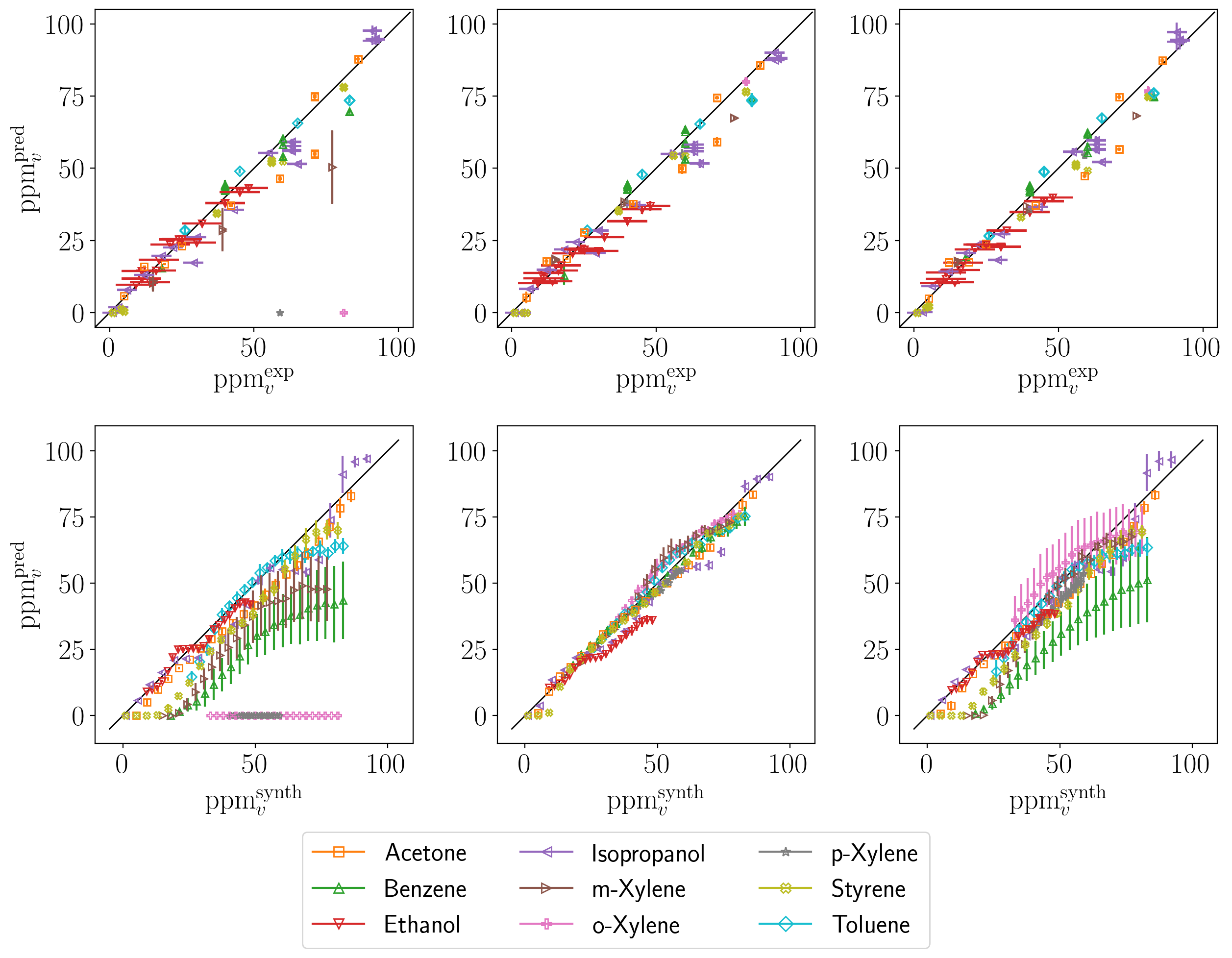}
    \caption{Predicted concentrations \emph{vs.} ground-truth values for the three discriminative NNs: \emph{basic} NN (left column),  \emph{synthetic} NN (central column), and \emph{oversample} NN (right column). The test is performed on the experimental dataset (upper row) and on synthetic spectra generated by the VAE (lower row).
    }
    \label{fig:regression-differences}
\end{figure}

To further assess the relative performance of the \emph{enhanced} and \emph{basic} NNs, we test them on synthetic spectra generated by the CVAE. This allows us to consider VOC concentrations that are not included in the experimental dataset. Yet, to address realistic levels, we uniformly sample them within the experimental ranges.
The concentration predictions shown in Fig. \ref{fig:regression-differences} demonstrate that, while both \emph{enhanced} NNs outperform the \emph{basic} model, the \emph{synthetic} NN is the most accurate. Quantitatively, the MSEs on the whole test sets are $2(1) 10^{2} \mathrm{ppm}^2$, $23(5) \mathrm{ppm}^2$, and  $1.0(2) 10^3 \mathrm{ppm}^2$, for the \emph{oversample}, \emph{synthetic}, and \emph{basic} NNs, respectively.

\section{Conclusions}
\label{secconclusions}

In this work, we created a quite large experimental dataset of IR absorption spectra of nine classes of VOCs in air, of high interest for indoor and outdoor air-quality monitoring. The dataset includes IR spectra at various concentrations, ranging from tens of ppm to a few ppm, ready to be used for automatic recognition studies.
On the basis of this experimental dataset, we developed an ML approach for the data-driven automatic identification and quantification of VOCs in air by analyzing IR absorption spectra.
A key advantage of CNNs consists in the inclusions of several useful spectral features con the classification and concentration prediction.
To overcome the limited number of examples in the experimental dataset, particularly for under-represented VOC classes, a generative NN in the form of a CVAE was trained to generate synthetic spectra. These synthetic spectra may correspond to concentrations not originally present in the experimental dataset, paving the way for data augmentation.
Importantly, our tests confirmed that the synthetic spectra have the same characteristic as the experimental measurements, making them comparably effective for the training of discriminative NNs.
Indeed, the CNNs trained with data augmentation show an increase in performance compared to models trained solely on the original experimental dataset. Notably, these CNNs are capable of accurately discriminating the identity of the different VOCs and reliably predicting the concentration of the identified VOCs in the air.
The proposed CNNs are ready to be incorporated in detecting devices for VOCs recognition.
To favor further investigations in this direction, we provide a large balanced dataset of synthetic spectra in the repository of Ref. \cite{DellaValle2025Dec}.



\section{Acknoweledgement}
We acknowledge financial support from the following projects:  PRIN 2022 PNRR project “Ultrasensitive detEction oF vocs and pAthogens” “P2022NMBAJ“; PRIN 2022 MUR project “Hybrid algorithms for quantum simulators” Project No. 2022H77XB7.

\bibliographystyle{elsarticle-num} 
\bibliography{bibliography}

@misc{leggeeuropea,
  title        = {Directive (EU) 2022/431 of the European Parliament and of the Council of 9 March 2022 amending Directive 2004/37/EC on the protection of workers from the risks related to exposure to carcinogens or mutagens at work},
  howpublished = {Official Journal of the European Union, L 88, 16 March 2022, pp. 1‑14},
  year         = {2022},
  note         = {Available at: \url{https://eur-lex.europa.eu/eli/dir/2022/431/oj/eng}}
}

@misc{gazzettaufficiale,
    title = {Decreto 10 ottobre 2008 — Disposizioni atte a regolamentare l’emissione di aldeide formica da pannelli a base di legno e manufatti con essi realizzati in ambienti di vita e soggiorno},
	howpublished = {Gazzetta Ufficiale della Repubblica Italiana, Serie Generale n. 288 del 10 dicembre 2008},
	year = {208},
	note = {Available at: \url{https://www.gazzettaufficiale.it/atto/serie_generale/caricaDettaglioAtto/originario?atto.dataPubblicazioneGazzetta=2008-12-10&atto.codiceRedazionale=08A09225&elenco30giorni=false}}
}

@article{Chowdhury2025Mar,
	author = {Chowdhury, M. A. Z. and Oehlschlaeger, M. A.},
	title = {{Artificial Intelligence in Gas Sensing: A Review}},
	journal = {ACS Sens.},
	volume = {10},
	number = {3},
	pages = {1538--1563},
	year = {2025},
	month = {mar},
	publisher = {American Chemical Society},
	doi = {10.1021/acssensors.4c02272}
}

@article{Tian2021Sep,
	author = {Tian, Linbo and Sun, Jiachen and Chang, Jun and Xia, Jinbao and Zhang, Zhifeng and Kolomenskii, Alexandre A. and Schuessler, Hans A. and Zhang, Sasa},
	title = {{Retrieval of gas concentrations in optical spectroscopy with deep learning}},
	journal = {Measurement},
	volume = {182},
	pages = {109739},
	year = {2021},
	month = sep,
	issn = {0263-2241},
	publisher = {Elsevier},
	doi = {10.1016/j.measurement.2021.109739}
}

@article{Chowdhury2024Mar,
	author = {Chowdhury, M. Arshad Zahangir and Oehlschlaeger, Matthew A.},
	title = {{Deep Learning for Gas Sensing via Infrared Spectroscopy}},
	journal = {Sensors},
	volume = {24},
	number = {6},
	pages = {1873},
	year = {2024},
	month = mar,
	issn = {1424-8220},
	publisher = {Multidisciplinary Digital Publishing Institute},
	doi = {10.3390/s24061873}
}

@article{Mancini2025Jun,
	author = {Mancini, Tiziana and Radica, Francesco and Mosesso, Lorenzo and Paolozzi, Maria Chiara and Macis, Salvatore and Marcelli, Augusto and Tamascelli, Stefano and Tranfo, Giovanna and Della Ventura, Giancarlo and Lupi, Stefano and D{'}Arco, Annalisa},
	title = {{Ultrahigh-sensitive and real-time detection of BTXs for occupational safety via infrared spectroscopy coupled with machine learning technique}},
	journal = {J. Environ. Chem. Eng.},
	volume = {13},
	number = {3},
	pages = {116833},
	year = {2025},
	month = jun,
	issn = {2213-3437},
	publisher = {Elsevier},
	doi = {10.1016/j.jece.2025.116833}
}

@article{Olive1964Aug,
	author = {Dunn, O. J.},
	title = {{Multiple Comparisons Using Rank Sums}},
	journal = {Technometrics},
	year = {1964},
	month = aug,
	issn = {1049-0181},
	publisher = {Taylor {\&} Francis Group},
	doi = {10.1080/00401706.1964.10490181}
}

@article{Terpilowski2019Apr,
	author = {Terpilowski, Maksim A.},
	title = {{scikit-posthocs: Pairwise multiple comparison tests in Python}},
	journal = {Journal of Open Source Software},
	volume = {4},
	number = {36},
	pages = {1169},
	year = {2019},
	month = apr,
	issn = {2475-9066},
	doi = {10.21105/joss.01169}
}

@article{Kruskal1952Dec,
	author = {Kruskal, W. H. and Wallis W. A.},
	title = {{Use of Ranks in One-Criterion Variance Analysis}},
	journal = {J. Am. Stat. Assoc.},
	year = {1952},
	month = dec,
	issn = {1048-3441},
	publisher = {Taylor {\&} Francis Group},
	doi = {10.1080/01621459.1952.10483441}
}

@ARTICLE{2020SciPy-NMeth,
  author  = {Virtanen, Pauli and Gommers, Ralf and Oliphant, Travis E. and
            Haberland, Matt and Reddy, Tyler and Cournapeau, David and
            Burovski, Evgeni and Peterson, Pearu and Weckesser, Warren and
            Bright, Jonathan and {van der Walt}, St{\'e}fan J. and
            Brett, Matthew and Wilson, Joshua and Millman, K. Jarrod and
            Mayorov, Nikolay and Nelson, Andrew R. J. and Jones, Eric and
            Kern, Robert and Larson, Eric and Carey, C J and
            Polat, {\.I}lhan and Feng, Yu and Moore, Eric W. and
            {VanderPlas}, Jake and Laxalde, Denis and Perktold, Josef and
            Cimrman, Robert and Henriksen, Ian and Quintero, E. A. and
            Harris, Charles R. and Archibald, Anne M. and
            Ribeiro, Ant{\^o}nio H. and Pedregosa, Fabian and
            {van Mulbregt}, Paul and {SciPy 1.0 Contributors}},
  title   = {{{SciPy} 1.0: Fundamental Algorithms for Scientific
            Computing in Python}},
  journal = {Nature Methods},
  year    = {2020},
  volume  = {17},
  pages   = {261--272},
  adsurl  = {https://rdcu.be/b08Wh},
  doi     = {10.1038/s41592-019-0686-2},
}

@article{Salthammer2016Feb,
	author = {Salthammer, T.},
	title = {{Very volatile organic compounds: an understudied class of indoor air pollutants}},
	journal = {Indoor Air},
	volume = {26},
	number = {1},
	pages = {25--38},
	year = {2016},
	month = feb,
	issn = {0905-6947},
	publisher = {John Wiley {\&} Sons, Ltd},
	doi = {10.1111/ina.12173}
}

@article{Lanyon2005,
	author = {Lanyon, Yvonne H. and Marrazza, G. and Tothill, I. and Mascini, M.},
	title = {{Benzene analysis in workplace air using an FIA-based bacterial biosensor.}},
	journal = {Biosens. Bioelectron.},
	year = {2005},
	doi = {10.1016/j.bios.2004.08.034}
}

@article{Rizk2018Nov,
	author = {Rizk, Malak and Guo, Fangfang and Verriele, Marie and Ward, Michael and Dusanter, Sebastien and Blond, Nad{\ifmmode\grave{e}\else\`{e}\fi}ge and Locoge, Nadine and Schoemaecker, Coralie},
	title = {{Impact of material emissions and sorption of volatile organic compounds on indoor air quality in a low energy building: Field measurements and modeling}},
	journal = {Indoor Air},
	volume = {28},
	number = {6},
	pages = {924--935},
	year = {2018},
	issn = {0905-6947},
	publisher = {John Wiley {\&} Sons, Ltd},
	doi = {10.1111/ina.12493}
}

@article{mishra2022,
	author = {Mishra, Puneet and Passos, D{\ifmmode\acute{a}\else\'{a}\fi}rio and Marini, Federico and Xu, Junli and Amigo, Jose M. and Gowen, Aoife A. and Jansen, Jeroen J. and Biancolillo, Alessandra and Roger, Jean Michel and Rutledge, Douglas N. and Nordon, Alison},
	title = {{Deep learning for near-infrared spectral data modelling: Hypes and benefits}},
	journal = {TrAC, Trends Anal. Chem.},
	volume = {157},
	pages = {116804},
	year = {2022},
	month = dec,
	issn = {0165-9936},
	publisher = {Elsevier},
	doi = {10.1016/j.trac.2022.116804}
}

@article{ho2019,
	author = {Ho, Chi-Sing and Jean, Neal and Hogan, Catherine A. and Blackmon, Lena and Jeffrey, Stefanie S. and Holodniy, Mark and Banaei, Niaz and Saleh, Amr A. E. and Ermon, Stefano and Dionne, Jennifer},
	title = {{Rapid identification of pathogenic bacteria using Raman spectroscopy and deep learning}},
	journal = {Nat. Commun.},
	volume = {10},
	number = {4927},
	pages = {1--8},
	year = {2019},
	month = oct,
	issn = {2041-1723},
	publisher = {Nature Publishing Group},
	doi = {10.1038/s41467-019-12898-9}
}

@incollection{riad2019,
	author = {Riad, Michael M. Y. R. and Sabry, Yasser M. and Khalil, Diaa},
	title = {{On the Detection of Volatile Organic Compounds (VOCs) Using Machine Learning and FTIR Spectroscopy for Air Quality Monitoring}},
	booktitle = {{2019 36th National Radio Science Conference (NRSC)}},
	journal = {Published in: 2019 36th National Radio Science Conference (NRSC)},
	pages = {16--18},
    year = {2019},
	publisher = {IEEE},
	doi = {10.1109/NRSC.2019.8734644}
}

@article{wang2021,
	author = {Wang, Ching-Yu and Ko, Tsung-Shun and Hsu, Cheng-Che},
	title = {{Interpreting convolutional neural network for real-time volatile organic compounds detection and classification using optical emission spectroscopy of plasma}},
	journal = {Anal. Chim. Acta},
	volume = {1179},
	pages = {338822},
	year = {2021},
	month = sep,
	issn = {0003-2670},
	publisher = {Elsevier},
	doi = {10.1016/j.aca.2021.338822}
}

@article{davaslioglu2018,
	author = {Davaslioglu, Kemal and Sagduyu, Yalin E.},
	title = {{Generative Adversarial Learning for Spectrum Sensing}},
	journal = {arXiv},
	year = {2018},
	month = apr,
	eprint = {1804.00709},
	doi = {10.48550/arXiv.1804.00709}
}

@article{chung2024,
	author = {Chung, Jihoon and Zhang, Junru and Saimon, Amirul Islam and Liu, Yang and Johnson, Blake N. and Kong, Zhenyu},
	title = {{Imbalanced spectral data analysis using data augmentation based on the generative adversarial network}},
	journal = {Sci. Rep.},
	volume = {14},
	number = {13230},
	pages = {1--15},
	year = {2024},
	month = jun,
	issn = {2045-2322},
	publisher = {Nature Publishing Group},
	doi = {10.1038/s41598-024-63285-4}
}

@article{schiemer2024,
	author = {Schiemer, Robin and R{\ifmmode\ddot{u}\else\"{u}\fi}dt, Matthias and Hubbuch, J{\ifmmode\ddot{u}\else\"{u}\fi}rgen},
	title = {{Generative data augmentation and automated optimization of convolutional neural networks for process monitoring}},
	journal = {Front. Bioeng. Biotechnol.},
	volume = {12},
	pages = {1228846},
	year = {2024},
	month = jan,
	issn = {2296-4185},
	publisher = {Frontiers},
	doi = {10.3389/fbioe.2024.1228846}
}

@article{darco2022,
	author = {D{'}Arco, Annalisa and Mancini, Tiziana and Paolozzi, Maria Chiara and Macis, Salvatore and Mosesso, Lorenzo and Marcelli, Augusto and Petrarca, Massimo and Radica, Francesco and Tranfo, Giovanna and Lupi, Stefano and Della Ventura, Giancarlo},
	title = {{High Sensitivity Monitoring of VOCs in Air through FTIR Spectroscopy Using a Multipass Gas Cell Setup}},
	journal = {Sensors},
	volume = {22},
	number = {15},
	pages = {5624},
	year = {2022},
	month = jul,
	issn = {1424-8220},
	publisher = {Multidisciplinary Digital Publishing Institute},
	doi = {10.3390/s22155624}
}

@article{VOCNet2022,
	author = {Chowdhury, M. Arshad Zahangir and Rice, Timothy E. and Oehlschlaeger, Matthew A.},
	title = {{VOC-Net: A Deep Learning Model for the Automated Classification of Rotational THz Spectra of Volatile Organic Compounds}},
	journal = {Appl. Sci.},
	volume = {12},
	number = {17},
	pages = {8447},
	year = {2022},
	month = aug,
	issn = {2076-3417},
	publisher = {Multidisciplinary Digital Publishing Institute},
	doi = {10.3390/app12178447}
}

@article{Kingma2022,
	author = {Kingma, Diederik P. and Welling, Max},
	title = {{Auto-Encoding Variational Bayes}},
	journal = {arXiv},
	year = {2013},
	month = dec,
	eprint = {1312.6114},
	doi = {10.48550/arXiv.1312.6114}
}

@article{higgins2018,
	author = {Higgins, Irina and Sonnerat, Nicolas and Matthey, Loic and Pal, Arka and Burgess, Christopher P. and Bosnjak, Matko and Shanahan, Murray and Botvinick, Matthew and Hassabis, Demis and Lerchner, Alexander},
	title = {{SCAN: Learning Hierarchical Compositional Visual Concepts}},
	journal = {arXiv},
	year = {2017},
	month = jul,
	eprint = {1707.03389},
	doi = {10.48550/arXiv.1707.03389}
}

@article{Sohn2015,
	author = {Sohn, Kihyuk and Lee, Honglak and Yan, Xinchen},
	title = {{Learning Structured Output Representation using Deep Conditional Generative Models}},
	journal = {Advances in Neural Information Processing Systems},
	volume = {28},
	year = {2015},
	note = {Available at: \url{https://papers.nips.cc/paper_files/paper/2015/hash/8d55a249e6baa5c06772297520da2051-Abstract.html} }
}

@article{Zeng2023Jun,
	author = {Zeng, Chunyan and Yan, Kang and Wang, Zhifeng and Yu, Yan and Xia, Shiyan and Zhao, Nan},
	title = {{Abs-CAM: a gradient optimization interpretable approach for explanation of convolutional neural networks}},
	journal = {SIViP.},
	volume = {17},
	number = {4},
	pages = {1069--1076},
	year = {2023},
	month = jun,
	issn = {1863-1711},
	publisher = {Springer London},
	doi = {10.1007/s11760-022-02313-0}
}

@article{McInnes2018Feb,
	author = {McInnes, Leland and Healy, John and Melville, James},
	title = {{UMAP: Uniform Manifold Approximation and Projection for Dimension Reduction}},
	journal = {arXiv},
	year = {2018},
	month = feb,
	eprint = {1802.03426},
	doi = {10.48550/arXiv.1802.03426}
}

@article{Galstyan2021,
	author = {Galstyan, Vardan and D’Arco, Annalisa and Di Fabrizio, Marta and Poli, Nicola and Lupi, Stefano and Comini, Elisabetta},
	title = {{Detection of volatile organic compounds: From chemical gas sensors to terahertz spectroscopy}},
	journal = {Reviews in Analytical Chemistry},
	year = {2021},
        volume= {40(1)},
	page= {33-57},
        doi={10.1515/revac-2021-0127}
}

@article{Radica2021,
	author = {Radica, Francesco and Della Ventura, Giancarlo and Malfatti, Luca and Cestelli Guidi, Mariangela and D'Arco, Annalisa and Grilli, Antonio and Marcelli, Augusto and Innocenzi, Plinio},
	title = {{Real-time quantitative detection of styrene in atmosphere in presence of other volatile-organic compounds using a portable device}},
	journal = {Talanta},
	year = {2021},
        volume= {233},
	page= {122510},
        doi={10.1016/j.talanta.2021.122510}
}

@article{Sicard2023,
	author = {Sicard, P. and Agathokleous, E. and Anenberg, S.C. and De Marco, A. and Paoletti, E. and Calatayud, V.},
	title = {{Trends in urban air pollution over the last two decades: a global perspective}},
	journal = {Sci. Total Environ.},
	year = {2023},
        volume= {858 (22)},
	page= {160064},
        doi={10.1016/j.scitotenv.2022.160064}
}

@article{DArco2022a,
	author = {D'Arco, A. and Rocco, D. and Piamonte Magboo, F. Jr and Moffa, C. and Dell Ventura, G. and Marcelli, A. and Palumbo, L. and Mattiello, L. and Lupi, S. and Petrarca, M.},
	title = {{Terahertz continuous wave spectroscopy: a portable advanced method for atmospheric gas sensing}},
	journal = {Optics Express},
	year = {2022},
        volume= {30 (11)},
	page= {19005--19016},
        doi={10.1364/OE.456022}
}

@article{Kullback1951Mar,
	author = {Kullback, S. and Leibler, R. A.},
	title = {{On Information and Sufficiency}},
	journal = {Ann. Math. Stat.},
	volume = {22},
	number = {1},
	pages = {79--86},
	year = {1951},
	month = mar,
	issn = {0003-4851},
	publisher = {Institute of Mathematical Statistics},
	doi = {10.1214/aoms/1177729694}
}

@article{Bouzerda2025,
	author = {Bouzerda, Outmane and Kane, Laura E. and Mellotte, Gregory S. and Ryan, Barbara M. and Maher, Stephen G. and Piot, Olivier and Meade, Aidan D.},
	title = {{Ex vivo spectroscopic characterisation of the biological activity of pancreatic cyst fluid}},
	journal = {Analyst},
	volume = {150},
	number = {14},
	pages = {3123--3136},
	year = {2025},
	publisher = {Royal Society of Chemistry},
	doi = {10.1039/D5AN00230C}
}

@article{Hotelling1933AnalysisOA,
  title={Analysis of a complex of statistical variables into principal components.},
  author={Harold Hotelling},
  journal={Journal of Educational Psychology},
  year={1933},
  volume={24},
  pages={498-520},
  url={https://api.semanticscholar.org/CorpusID:144828484}
}

@article{DellaValle2025Dec,
	author = {Della Valle, Andrea and D'Arco, Annalisa and Mancini, Tiziana and Mosetti, Rosanna and Paolozzi, Maria Chiara and Lupi, Stefano and Pilati, Sebastiano and Perali, Andrea},
	title = {{Synthetic dataset from the manuscript "Deep learning recognition and analysis of Volatile Organic Compounds based on experimental and synthetic infrared absorption spectra"}},
	journal = {Zenodo},
	year = {2025},
	month = dec,
	doi = {10.5281/zenodo.17709018}
}

@article{Li2021Jan,
	author = {Li, Adela Jing and Pal, Vineet Kumar and Kannan, Kurunthachalam},
	title = {{A review of environmental occurrence, toxicity, biotransformation and biomonitoring of volatile organic compounds}},
	journal = {Environ. Chem. Ecotoxicol.},
	volume = {3},
	pages = {91--116},
	year = {2021},
	month = jan,
	issn = {2590-1826},
	publisher = {Elsevier},
	doi = {10.1016/j.enceco.2021.01.001}
}

@article{Lamm2009Dec,
	author = {Lamm, Steven H. and Engel, Arnold and Joshi, Kiran P. and Byrd, Daniel M. and Chen, Rusan},
	title = {{Chronic myelogenous leukemia and benzene exposure: A systematic review and meta-analysis of the case{\textendash}control literature}},
	journal = {Chem.-Biol. Interact.},
	volume = {182},
	number = {2},
	pages = {93--97},
	year = {2009},
	month = dec,
	issn = {0009-2797},
	publisher = {Elsevier},
	doi = {10.1016/j.cbi.2009.08.010}
}

@article{Mazzatenta2015Apr,
	author = {Mazzatenta, Andrea and Pokorski, Mieczyslaw and Sartucci, Ferdinando and Domenici, Luciano and Di Giulio, Camillo},
	title = {{Volatile organic compounds (VOCs) fingerprint of Alzheimer's disease}},
	journal = {Respir. Physiol. Neurobiol.},
	volume = {209},
	pages = {81--84},
	year = {2015},
	month = apr,
	issn = {1569-9048},
	publisher = {Elsevier},
	doi = {10.1016/j.resp.2014.10.001}
}

@article{Talibov2018Aug,
	author = {Talibov, Madar and Sormunen, Jorma and Hansen, Johnni and Kjaerheim, Kristina and Martinsen, Jan-Ivar and Sparen, Per and Tryggvadottir, Laufey and Weiderpass, Elisabete and Pukkala, Eero},
	title = {{Benzene exposure at workplace and risk of colorectal cancer in four Nordic countries}},
	journal = {Cancer Epidemiology},
	volume = {55},
	pages = {156--161},
	year = {2018},
	month = aug,
	issn = {1877-7821},
	publisher = {Elsevier},
	doi = {10.1016/j.canep.2018.06.011}
}


\end{document}